\documentclass[lettersize,journal]{IEEEtran}
\usepackage{amsmath,amsfonts}
\usepackage{subcaption}
\usepackage{array}
\usepackage{soul} 
\usepackage{booktabs}
\usepackage{subcaption}
\usepackage{amsmath}
\usepackage{hyperref}
\usepackage{cleveref}
\usepackage{textcomp}
\usepackage{stfloats}
\usepackage{url}
\usepackage{verbatim}
\usepackage{graphicx}
\usepackage[table]{xcolor}
\usepackage[ruled,linesnumbered]{algorithm2e}
\usepackage{threeparttable}
\usepackage{multirow}
\usepackage{multicol}
\usepackage{array}
\usepackage{tabularx}
\usepackage{cite}
\usepackage{etoolbox}
\newcommand{\etal}{\textit{et al.}}

\begin{document}

\title{BENet: A Cross-domain Robust Network for Detecting Face Forgeries via Bias Expansion and Latent-space Attention}

\author{\IEEEauthorblockN{Weihua Liu,
Jianhua Qiu,
Said Boumaraf\thanks{Corresponding author: Said Boumaraf (email: said.boumaraf@yahoo.com).}, 
Chaochao lin,
Pan liyuan,
Lin Li,
Mohammed Bennamoun,and
Naoufel Werghi}
\thanks{\textit{Weihua Liu, Jianhua Qui, and Said Boumaraf contributed equally to this work.}}}

\markboth{ArXiv Preprint}%
{Shell \MakeLowercase{\textit{et al.}}: A Sample Article Using IEEEtran.cls for IEEE Journals}


\maketitle

\begin{abstract}
In response to the growing threat of deepfake technology, we introduce BENet, a Cross-Domain Robust Bias Expansion Network. BENet enhances the detection of fake faces by addressing limitations in current detectors related to variations across different types of fake face generation techniques, where ``cross-domain" refers to the diverse range of these deepfakes, each considered a separate domain. BENet's core feature is a bias expansion module based on autoencoders. This module maintains genuine facial features while enhancing differences in fake reconstructions, creating a reliable bias for detecting fake faces across various deepfake domains. We also introduce a Latent-Space Attention (LSA) module to capture inconsistencies related to fake faces at different scales, ensuring robust defense against advanced deepfake techniques. The enriched LSA feature maps are multiplied with the expanded bias to create a versatile feature space optimized for subtle forgeries detection. To improve its ability to detect fake faces from unknown sources, BENet integrates a cross-domain detector module that enhances recognition accuracy by verifying the facial domain during inference. We train our network end-to-end with a novel bias expansion loss, adopted for the first time, in face forgery detection. Extensive experiments covering both intra and cross-dataset demonstrate BENet's superiority over current state-of-the-art solutions.
\end{abstract}

\begin{IEEEkeywords}
Face forgery detection, cross-domain detector, latent-space attention, bias expansion, deep learning. 
\end{IEEEkeywords}

\section{Introduction}
\label{sec:intro}
Recent advancements in Deepfake generation techniques have been remarkably impressive, producing convincingly fake facial images or videos. Exploiting these methods, an attacker could create deceptive news, defame public figures, or compromise security systems, leading to substantial risks with broad social and security consequences~\cite{Zhang2022, Nguyen2022, Seow2022}. Face forgery detection has arisen in response to these manipulated visuals, concentrating on the discriminative task of identifying forged regions through visual analysis.

Current face forgery detection methods~\cite{li2020identification, he2019detection, mccloskey2019detecting, guarnera2020deepfake, zhou2017two, rossler2019faceforensics++, li2020face, qian2020thinking, li2021frequency, yang2022deepfake, cao2022end} primarily focus on identifying forgery clues left by generative models, e.g., GANs~\cite{goodfellow2020generative}. They model detection as a binary classification task using a backbone network for learning global facial image representations, followed by a binary classifier to distinguish between real and fake images. However, forgers often manipulate faces to hide forgery clues, making detection challenging. Fortunately, deepfake methods typically struggle to mimic real faces' unique statistical pixel distribution, leading to noticeable inconsistencies between tampered and authentic regions~\cite{yu2019attributing}. This disparity is a key factor in developing proactive methods to highlight forgery clues efficiently. 

On the other hand, advancements in deepfake technology have led to more sophisticated forgeries, blurring the line between real and fake content. As forgeries improve, the telltale signs become more nuanced and localized, rendering the global feature-based methods less effective. Also, the expanding spectrum of deepfake techniques challenges the cross-domain robustness of face forgery detection models,  where \textbf{``cross-domain”} refers to the capability of these models to accurately identify fake faces generated by a variety of different deepfakes, each considered a separate domain. Even though existing solutions for detecting forgeries perform well on examples that share the same statistical distribution as the training data, they expect significant drops in effectiveness when the conditions deviate from this strict similarity~\cite{li2020face, nguyen2019multi, chai2020makes}.

With the above considerations in mind, in this paper, we propose BENet, a Cross-Domain Robust Bias Expansion Network, designed to augment face forgery detection capabilities. BENet incorporates a bias expansion module-based autoencoder for modeling the distributions of both real and fake faces. Specifically, while real faces exhibit consistent features, leading to largely invariant reconstructions by BENet, the network distinctly intensifies forgery clues when applied to fake faces and expands the bias against them. This approach establishes a dependable bias expansion for efficient forgery clues detection.

Inspired by breakthroughs in Transformer attention mechanisms~\cite{vaswani2017attention, dosovitskiy2020image}, we propose a novel Latent-space Attention (LSA) module that emphasizes the forgery-related inconsistencies by capturing the latent feature variances across multiple scales between the encoder and decoder. This multi-scale LSA approach not only captures fine-grained manipulations but also adapts to the holistic context of the face, ensuring a robust defense against even the most sophisticated forgeries.


Furthermore, we incorporate a cross-domain detector module to strengthen BENet’s capability in identifying unfamiliar cross-domain deepfake attacks. Traditionally, An open-set classifier should reject unknown samples while maintaining high accuracy on known ones~\cite{scheirer2012toward, yoshihashi2019classification}. Intuitively, we formulate a \textit{Bias-against-Threshold} filter to catch images that are significantly different (\textit{unknown fakes}) from what the model was trained to recognize (\textit{known ``real and fake" images}). During training, BENet learns to differentiate real from fake faces by fitting a \textit{posterior probability distribution} to training data features. In inference, it detects fakes by measuring how much an image's bias deviates from this learned distribution. Thus, input samples that exceed a specific bias threshold are automatically classified as \textit{fake}, ensuring a robust detection across various deepfake domains (see \cref{sec:3.4} for more details). This method enables BENet to learn a feature space that effectively distinguishes real and fake samples, leading to higher accuracy and better generalization when compared to other methods. Our main contributions can be outlined as follows:




\begin{itemize}
  \item We present BENet, a specifically designed network for effective detection of face forgery. BENet incorporates an autoencoder with a bias expansion module, enhancing forgery clues identification by amplifying the bias between real and fake reconstructions.
  
  \item We introduce the innovative Latent-Space Attention (LSA) module within BENet, enabling the capture of latent feature variances across multiple scales between the encoder and decoder networks, for getting a comprehensive feature representation of fine-grained forgery clues.
  
  \item We propose an original cross-domain detector module to boost BENet generalization on unseen manipulations. This module is activated during inference to strengthen BENet’s capability to identify unfamiliar cross-domain deepfake attacks.
  
   \item Through comprehensive experiments on diverse state-of-the-art benchmarks, we demonstrate the superior performance of BENet compared to its counterparts in intra and cross-dataset evaluations.


\end{itemize}

\section{Related works}
\label{sec:related}

\subsection{Face Forgery Detection}
Earlier approaches to face forgery detection focused on hand-crafted features, particularly by examining inconsistencies in the color space~\cite{li2020identification, he2019detection, mccloskey2019detecting, guarnera2020deepfake}. With recent strides in deep learning, CNNs have emerged as powerful tools, demonstrating significant performance gains~\cite{zhou2017two, rossler2019faceforensics++, li2020face, qian2020thinking, li2021frequency, yang2022deepfake, cao2022end}.  
For instance, Zhou \etal~\cite{zhou2017two} combined face classification with noise residual recognition, while Rossler \etal~\cite{rossler2019faceforensics++} used an Xception model for binary classification based on features extracted from cropped facial images. Qian \etal~\cite{qian2020thinking} and Li \etal~\cite{li2021frequency} developed models focusing on frequency aspects to distinguish authentic and counterfeit faces. To uncover correlated input features and identifying anomalies within datasets, \textit{autoencoders} have become instrumental in the last few years~\cite{nguyen2019multi, chakraborty2019integration,chen2020unsupervised,dai2019multilayer,sarvari2021unsupervised, pimentel2020deep,akhriev2019deep}. In the context of deepfake limitations in replicating genuine statistical pixel distribution, some research has incorporated autoencoders in this effort. For instance, Nguyen \etal~\cite{nguyen2019multi} designed a CNN with a multi-task learning approach to detect and locate manipulated regions within images. Liu \etal~\cite{liu2023fedforgery} integrated a variational autoencoder into an innovative generalized residual federated learning scheme named FedForgery to learn discriminative residual feature maps. In a similar vein, Khalid \etal~\cite{khalid2020oc} introduced OCFakeDect, utilizing a one-class variational autoencoder (VAE) trained solely on authentic facial images.
 Cao \etal~\cite{cao2022end} detect differences between genuine and fake faces via a bipartite graph-based fusion of the autoencoder latent  space and the decoder feature maps. \textit{As opposed to our approach, it is important to highlight that these methods apply autoencoders \textit{\textbf{exclusively}} to real facial images, bypassing direct modeling of fake sample distribution. Thus, there is no evidence that the learned representations will be generalizable.}
\subsection{Attention for Face Forgery Detection}
Recently, numerous studies based on attention mechanisms have been proposed for face forgery detection~\cite{yadav2023uncovering, dang2020detection, zhao2021multi}, for the purpose of spotting forgeries and detecting subtle discrepancies more effectively. For example, Yadav \etal ~\cite{yadav2023uncovering} used the `Local Attentional Tamper Trace Extractor' (LATTE) to extract forgery traces at the block level, while the `Global Attentional Tamper Trace Extracter' (GATTE) to aggregate tampering traces on a broader scale.  Transformers~\cite{dosovitskiy2020image} have recently been adapted for deepfake detection in models such as FTCN-TT~\cite{zheng2021exploring}, ICT~\cite{dong2022protecting}, and AUNet~\cite{bai2023aunet}. Although these methods achieve promising results in
the intra-domain, their performance significantly drops in the cross-domain evaluation. To cope with these, we: 1) Propose for the first time, an original aggregation strategy that integrates a {\it latent-space attention mechanism} (\cref{sec:3.2}) for emphasizing the forgery-related inconsistencies and capturing the latent feature variances across multiple scales between the encoder and decoder, 2) We introduce a {\it cross-domain detector} (\cref{sec:3.4}), which leverages the stability of the distribution of face features, compared to the fake face’s counterpart, to best counter cross-domain manipulations. We showcase that these newly proposed modules ensure a robust detection of forgeries across various domains.
\begin{figure*}[t]
  \centering
   \includegraphics[width=\linewidth]{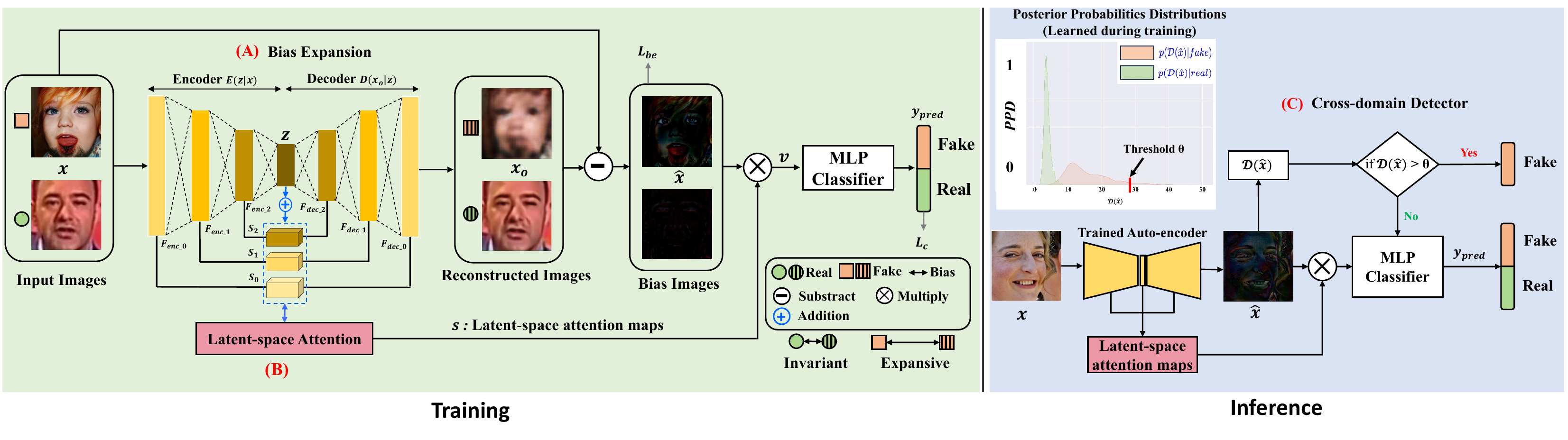}
   \caption{Overview of BENet Architecture. Three components play an important role in BENet: \textbf{(A)} a Bias expansion module for processing input images and amplifying forgery clues,  \textbf{(B)} A Latent-space attention (LSA) module for capturing the latent feature variances across multiple scales between the encoder and decoder, and \textbf{(C)} a Cross-domain detector module for enhancing defense against unknown attacks. The learning process of BENet is optimized end-to-end with a newly designed bias expansion loss (\cref{sec: 3.3}).
   }
   \label{fig:2}
\end{figure*}

\section{Methodology}
For enhanced intra-domain and cross-domain face forgery detection, we propose BENet, which consists of three main components, namely Bias expansion, latent-space attention, and cross-domain detector, as shown in \cref{fig:2}. \textbf{(A)} Input images $x$ (real  or fake) are first processed by a bias expansion module (i.e., autoencoder), yielding reconstructed images $x_o=D(E(x))$. The resultant bias images $\hat{x}$ are derived by subtracting $x$ from $x_o$. \textbf{(B)} A multi-scale Latent-space Attention (LSA) module further emphasizes the forgery-related inconsistencies by capturing the latent feature variances between the encoder and decoder. The enriched LSA feature maps are then multiplied with the expanded bias images, resulting in a multi-faceted feature space used as input for a multi-layer perceptron (MLP) classifier for binary classification. BENet learning is optimized end-to-end by a newly designed contrastive loss. \textbf{(C)} A thresholded cross-domain detector is introduced further to improve the recognition accuracy during inference through facial domain verification. 
We detailed the BENet's components in the following subsections.

\subsection{Bias Expansion Module } 
\label{sec: 3.3}
In order to expand the bias between real and fake reconstructions, we employ an autoencoder (\cref{fig:2} (A)) to obtain reconstructed images $x_o$, which amplifies the deepfake clues of input images $x$. The reconstructed images $x_o$ are defined as:
\begin{equation}
  x_o=AE(x)
  \label{eq:1}
\end{equation}
where $AE(\cdot)$ represents the reconstruction process of the autoencoder. Then, we calculate the bias images $\hat{x}$  by subtraction, which are denoted as:
\begin{equation}
  \hat{x}=|x-x_o|
  \label{eq:2}
\end{equation}


The bias images are the pixel-level difference between the input images and the reconstructed versions, indicating deepfake clues. We aim to expand deepfake bias while retaining the reconstructed real faces invariant. This is consistent with the idea of contrastive loss~\cite{hadsell2006dimensionality}. Therefore, we define bias expansion loss $L_{be}$ as follows: 
\begin{equation}
  L_{be}=L_1+L_2+L_3
  \label{eq:3}
\end{equation}

\begin{figure}
 \centering
 \begin{subfigure}{0.22\linewidth}
   \includegraphics[width=\linewidth]{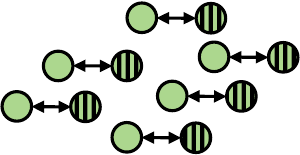}
   \caption{}
   \label{fig:3a}
 \end{subfigure}
 \hfill
 \begin{subfigure}{0.26\linewidth}
    \includegraphics[width=\linewidth]{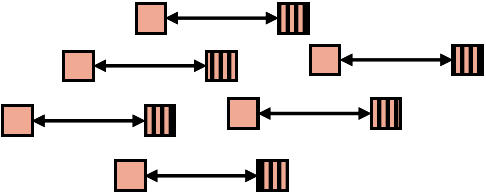}
   \caption{}
   \label{fig:3b}
  \end{subfigure}
 \hfill
  \begin{subfigure}{0.5\linewidth}
    \includegraphics[width=\linewidth]{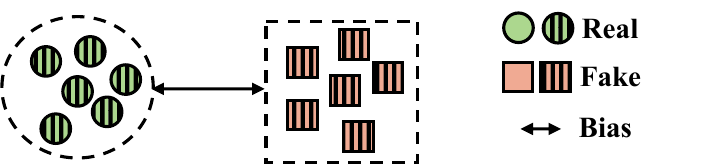}
    \caption{}
   \label{fig:3c}
  \end{subfigure}
 \caption{Illustration of bias expansion, showcasing the role of $L_1$, $L_2$, and $L_3$ . (a) $L_1$  maintains consistency between input real faces and their reconstructions. (b) $L_2$ expands the difference between input fake faces and their reconstructions. (c) $L_3$ increases the difference between real and fake faces biases.}
 \label{fig:Lbc}
\end{figure}

Here,
\begin{equation}
  L_1=\frac{1}{N} \sum_i^N {(1-y_i)||\hat{x}_i ||_2^2}
  \label{eq:4}
\end{equation}

\begin{equation}
  L_2=-\frac{1}{N}\sum_i^N {y_i \max(m-||\hat{x}_i ||_2,0)^2}
  \label{eq:5}
\end{equation}

\begin{equation}
  L_3=\frac{1}{N}\sum_i^N \frac{-1}{M}\sum_{j\neq i,y_i=y_j}^M\log\frac{\exp(\hat{x}_i\cdot\hat{x}_j)}{\sum_{k,k\neq i}^{N-1}\exp(\hat{x}_i\cdot\hat{x}_k)}
  \label{eq:6}
\end{equation}
$N$ is the number of samples from a batch, $y_i$ is the label of input image $x_i$, $M$ is the number of samples with $y_i=y_j$ from a batch, and $m$ is a margin parameter ensuring a minimum distance between original and reconstructed fake faces. BENet's bias expansion loss strengthens its ability to detect deepfake clues in fake faces. For real faces, $L_{be}$ aims to align reconstructed faces with originals by minimizing bias squared $\hat{x}_i^2$ within real face samples in $L_1$. In contrast, $L_2$ increases the disparity between original and reconstructed fake faces by maximizing this bias. Furthermore, we expand the differences between real and fake faces through $L_3$. The effect of each component in $L_{be}$ is illustrated in Figure \cref{fig:Lbc}. The synergistic aggregation of these three functions in BENet results in high sensitivity to even the slightest discrepancies introduced by facial forgery, enabling it to effectively identify fake faces. To the best of our knowledge, we are the first to adopt this newly designed bias expansion loss function for facial forgery detection.



\subsection{Latent-space Attention Module}  
\label{sec:3.2}

To achieve better representation of forged clues, we incorporate a novel Latent-Space Attention (LSA) module (\cref{fig:2} (B)) that emphasizes the forgery-related inconsistencies by capturing the latent feature variances across multiple scales between the encoder and decoder. Let $z$ denote the latent-space representation at the intermediary step. The computation of the autoencoder is as follows:

\begin{equation}
z=E(z|x)
  \label{eq:7}
\end{equation}
\begin{equation}
x_o=D(x_o |z)
  \label{eq:8}
\end{equation}

\begin{figure}[t]
   \centering
   \includegraphics[width=1\linewidth]{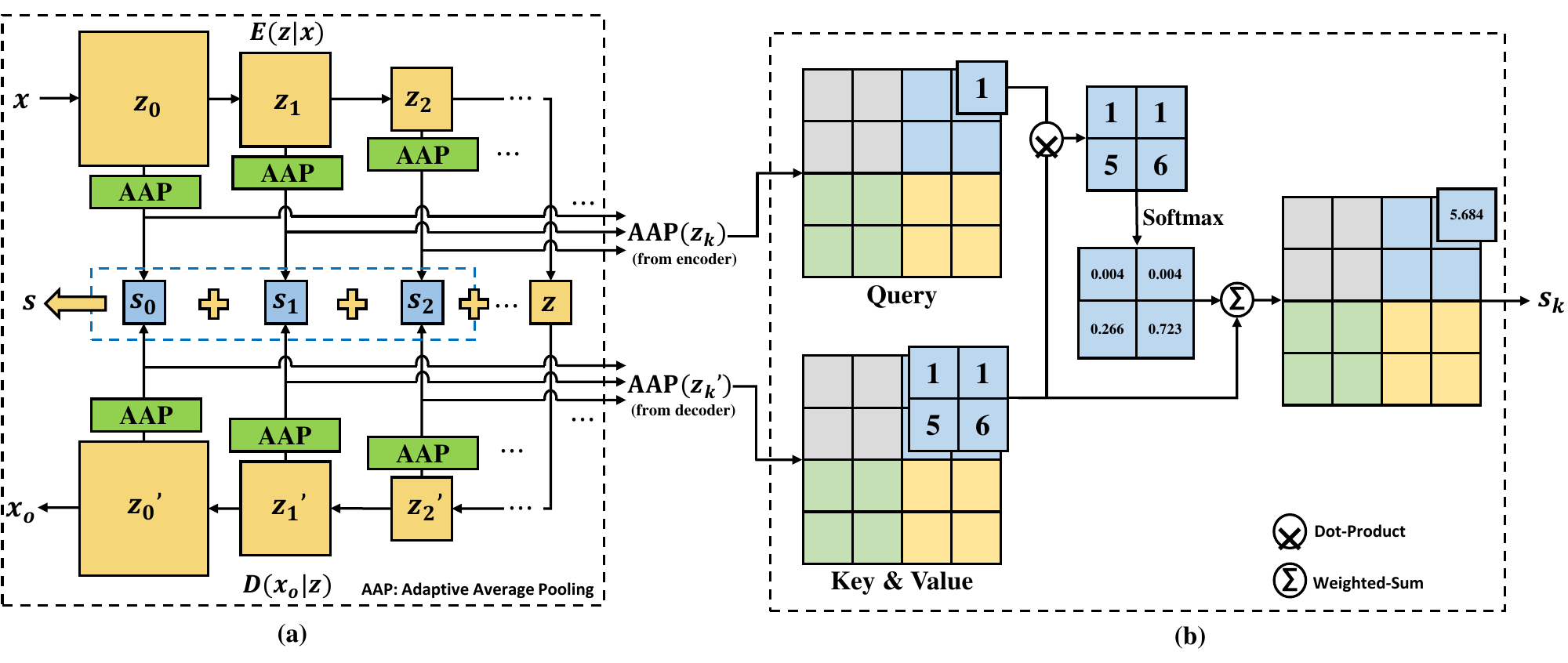}
   \caption{Overview of the LSA module. (a) The functional workflow of the LSA module; (b) The process for calculating latent-space attention maps (Per-channel calculation for simplicity).}
   \label{fig:3}
\end{figure}

In this context, $E(\cdot)$ and $D(\cdot)$ denote the encoding and decoding functions. The encoder's multi-scale latent-space representations are denoted as $z_0, z_1, z_2, \cdots, z_n$, and the decoder's corresponding representations are $z_0', z_1', z_2', \cdots, z_n'$. The LSA module employs Adaptive Average Pooling (AAP) to resize the multi-scale latent-space features from both the encoder and decoder to the dimensions of $z$. AAP is crucial for its flexibility in handling various input sizes and its efficiency in merging global spatial information from different scales. It dynamically adjusts pooling regions to produce a fixed-size output, preserving spatial information.
The process of generating latent-space attention maps is represented by $\mathrm{LSA}(\cdot,\cdot)$. These maps are computed for each feature map level. The final latent-space attention maps, labeled as $s$, result from aggregating these maps with the latent-space features $z$ across multiple scales, as shown in \cref{fig:3} (a). The equation for this calculation process is as follows:
\begin{equation}
s=\sum_{k=0}^n\mathrm{LSA}[\mathrm{AAP}(z_k),\mathrm{AAP}(z_k')]+z=\sum_{k=0}^n s_k + z
  \label{eq:9}
\end{equation}
Finally, The enriched LSA feature maps are then synergized (i.e., multiplied) with the expanded bias images $\hat{x}$, resulting in a multi-faceted feature maps $v$. These maps, specifically refined for subtle forgeries, are then fed as input into the classifier for face forgery detection. The formulation of the feature maps $v$ is defined as follows:

\begin{equation}
v=s\times\hat{x}
  \label{eq:10}
\end{equation}

Unlike global attention ~\cite{dosovitskiy2020image,zhao2020exploring}, which generally considers global relationships within the image, LSA specifically targets the variational relationships of latent-space features during reconstruction. It addresses forged details loss by focusing on subtle and localized patterns in small, face-aligned blocks, learning \textit{semantically} meaningful features and pinpointing key forgery clues with less computational overhead and improved detection efficiency.




To achieve this, we define $\mathrm{AAP}(z_k)$ as queries and $\mathrm{AAP}(z_k')$ as keys and values, corresponding to the encoded or decoded latent-space features at scale $k$. To integrate deepfake clues from model-generated inconsistencies, we break down these queries, keys, and values into $P\times P$ patches, as shown in \cref{fig:3} (b). A value $\beta\in\mathbb{R}$ in the latent-space attention maps $s_k$ is computed using the corresponding query value $\alpha\in\mathbb{R}$ and its associated patch $Z\in\mathrm{AAP}(z_k')$. This involves multiplying $\alpha$ by the key matrix patch $Z$, producing a $P\times P$ size matrix. The values within this matrix are then subjected to the softmax function for activation. The softmax output, a weighted sum of patch $Z$ from the value matrix, gives the final value $\beta$ in the latent-space attention maps, which is expressed as follows:

\begin{equation}
\beta=\mathrm{softmax}(\alpha Z)\cdot Z
  \label{eq:11}
\end{equation}

Through this computation of multi-scale LSA, BENet effectively focuses on the discrepancies in latent-space feature maps between the encoder and decoder. It not only captures fine-grained manipulations but also adapts to the holistic context of the input face, ensuring a robust defense against even the most sophisticated deepfake techniques.

\subsection{Total Loss}
The final comprehensive representation $v$,  derived from Equation~\ref{eq:10}, serves as the input to a binary multi-layer perceptron (MLP) classifier to distinguish real from fake. We use the \textit{cross entropy loss} $L_c$ as loss function: 

\begin{equation}
L_c=-\frac{1}{N} \sum_{i=1}^N [y_i\log p_i +(1-y_i)\log (1-p_i)]
  \label{eq:12}
\end{equation}
where $N$ is the number of samples from a batch, $y$ is the label, and $p$ is the predicted probability. 

By combining the bias expansion loss and the cross-entropy loss, the total loss of BENet is defined as:
\begin{equation}
L=\lambda L_c+(1-\lambda )L_{be}
  \label{eq:13}
\end{equation}
where $L_c$ denotes the objective of face forgery detection, and  $\lambda$ is a hyper-parameter to balance $L_c$ with the bias expansion loss $L_{be}$.  
\subsection{Cross-domain Detector Module} 
\label{sec:3.4}
In contrast to the stable and localized feature distributions observed in real faces, fake facial features exhibit a long-tail distribution. This discrepancy arises from various factors, encompassing the diverse and unpredictable nature of forgery techniques and the underlying intent of image manipulation. This phenomenon becomes evident when examining the posterior probabilities distributions $p(\mathcal{D}(\hat{x})|real)$ and
$ p(\mathcal{D}(\hat{x})| fake)$, depicted in \cref{fig:2}, where $\mathcal{D}(\hat{x})$  represents the average per-pixel discrepancy across the bias image, $\hat{x}$, defined as
$\mathcal{D}(\hat{x})=\frac{1}{N}\sum_{i=1}^N \hat{x}_i$, and $N$ is the number of pixels in $\hat{x}$.
Building upon this observation, we integrated a cross-domain detector module into BENet to identify fake instances that deviate from established distribution patterns, specifically those seen in the training data. Given the long-tail distribution characterizing fake samples, instances resulting from unknown image manipulations are expected to have a bias pixel discrepancy $\mathcal{D}(\hat{x})$ skewed towards the far right of the known distributions. Following this rationale, we designate an instance as a potentially \textit{unknown cross-domain fake} face if its bias pixel discrepancy exceeds a threshold, denoted as theta. We Determine theta by calculating $\mathcal{D}(\hat{x})$  for all training samples and selecting the threshold for which 95\% of training samples fall below it (see \cref{fig:2}).
\section{The Prediction Algorithm}
The Algorithm 1 outlines the prediction process that embodies the proposed methodology for detecting face forgery.

\begin{algorithm}
\caption{Prediction procedure for face forgery detection.}
\label{alg:algorithm}
\KwIn{Face image $x$ and threshold $\theta$ for bias}
\KwOut{Prediction label $y_{final}$}
Reconstructed image $x_o\leftarrow$AE$(x)$\;

Obtain bias image $\hat{x}\leftarrow|x-x_o|$\;

Obtain final latent-space attention maps $s$ from LSA module\;

Obtain feature maps $v=s\times\hat{x}$\;

Prediction probabilities, $p_y\leftarrow$Classifier$(v)$\;

Predict known label, $y_{pred}\leftarrow$argmax$(p_y)$\;

Calculate average per-pixel discrepancy $\mathcal{D}(\hat{x})$

\eIf{ $\mathcal{D}(\hat{x})$ $>$ $\theta$}{predict $x$ as unknown, $y_{final} \leftarrow $ fake\;}{predict $x$ as known, $y_{final} \leftarrow y_{pred}$\;}
\end{algorithm}

\section{Experiments}

\subsection{Experimental Setup}
\subsubsection{Datasets} We evaluate BENet and its counterparts on FaceForensics++ (FF++)~\cite{rossler2019faceforensics++}, Celeb-DF~\cite{li2020celeb}, Diverse Fake Face Dataset (DFFD)~\cite{dang2020detection}, and DeepFake Detection Challenge Dataset (DFDC)~\cite{dolhansky2020deepfake}. The FF++ dataset has 1,000 real videos from YouTube and 4,000 corresponding to four face manipulation methods: Deepfakes(DF)~\cite{Deepfakes}, FaceSwap(FS)~\cite{Faceswap}, Face2Face(F2F)~\cite{thies2016face2face}, and NeuralTextures(NT)~\cite{thies2019deferred}. The celeb-DF dataset contains 590 real videos and 5,639 Deepfake videos created using the same synthesis algorithm.  DFFD adopts the images from FFHQ~\cite{karras2019style} and CelebA~\cite{liu2015deep} datasets as the source subset and synthesizes forged images using various Deepfake generation methods. DFDC is a part of the DeepFake detection challenge, which has 1,131 original videos and 4,113 fake videos.


\subsubsection{Evaluation Metrics} To evaluate our proposed method, we report the most commonly used metrics in contemporary research, including accuracy (Acc) and area under the receiver operating characteristic curve (AUC). 

\subsubsection{Implementation Details} In our experiments, we used the dlib toolkit~\cite{dlib09} (version 19.24.0), renowned for its robust face recognition features. Dlib, a modern C++ toolkit, excels in machine learning algorithms and is particularly effective in computer vision tasks, including facial landmark detection and face recognition. Hence, we exploited these capabilities to accurately identify facial key points, facilitating the cropping and alignment of faces. These faces were then resized to a uniform 224×224 pixel size, forming the standardized input for our face forgery detection model, BENet. Note that this methodology is applied for all the three datasets, including FF++, Celeb-DF, and DFDC. For the DFFD~\cite{dang2020detection} dataset, which covers a diverse array of facial forgeries, the source images are directly adopted from the FFHQ~\cite{karras2019style} and CelebA ~\cite{liu2015deep} datasets. The DFFD provides \textit{pre-cropped faces} where the images are expertly transformed using various Deepfake generation methods. For all datasets, we adhere to the established train, validation, and test splits as originally defined and provided with each dataset. To augment our data, we applied random erasure and horizontal flipping. The network is trained using a batch size of 8, employing the Adam optimizer with an initial learning rate of 2e-4 and a weight decay of 1e-5. Note that the parameter $\lambda$ in BENet's objective formulation is set to 0.5 based on empirical determination.

We implemented our BENet and the other comparative variants using PyTorch 1.7.1, leveraging the open-source Xception~\cite{chollet2017xception} codebase. Our experiments were conducted using a Tesla NVIDIA A800-SXM4-80GB GPU and an Intel(R) Xeon(R) Platinum 8358 CPU operating at 2.60GHz.


\subsection{Experimental Results}
\subsubsection{Intra-evaluation} To evaluate the effectiveness and robustness of BENet, we conduct comprehensive comparison experiments with several state-of-the-art methods. Specifically, we experiment with the most challenging FF++ low-quality (LQ) dataset. This dataset is known for its complexity, which poses significant difficulties for many current methods in the field. Often, these existing approaches encounter struggles when dealing with the FF++ LQ, resulting in poor performance. As shown in \cref{tab:intra}, BENet's intra-evaluation performance outperforms other contemporary models in Acc/AUC with fair scores of 96.83/98.72(\%), 99.23/99.98(\%), 98.96/99.93(\%), and 90.43/96.38(\%) on FF++, Celeb-DF, DFFD, and DFDC, respectively. Particularly, we achieve a state-of-the-art AUC score of 98.72\% on FF++(LQ). 
Notably, in comparison to the approach in~\cite{zhao2021multi}, which employed a multi-attentional network concentrating on local features and textural details for addressing subtle forgeries, our method exhibits a remarkable improvement. This substantial performance boost highlights the efficacy of the introduced multi-scale attention mechanism applied to auto-encoder reconstructions.
Overall, BENet shows consistent robustness across various complex scenarios and different datasets.

\begin{table*}
  \centering
  \begin{threeparttable}
  \begin{tabularx}{\textwidth}{XXXXXXXXXX}
    \toprule
   \multicolumn{2}{c}{\multirow{2}{*}{Methods}} & \multicolumn{2}{c}{FF++ (LQ)} & \multicolumn{2}{c}{Celeb-DF} & \multicolumn{2}{c}{DFFD} & \multicolumn{2}{c}{DFDC} \\ 
   \cline{3-10}
   \multicolumn{2}{c}{\multirow{2}{*}{}} & Acc(\%) & AUC(\%)  & Acc(\%)  & AUC(\%)  & Acc(\%) & AUC(\%)  & Acc(\%)  & AUC(\%) \\
    \midrule
    \multicolumn{2}{l}{Multi-task \cite{nguyen2019multi}} & 81.30$^\ddag$ & 75.59$^\ddag$ & \ \ \  \textendash  & \ \ \ \textendash\  &  \ \ \ \textendash\  & \ \ \ \textendash\  & \ \ \ \textendash\  & \ \ \ \textendash\  \\
    \multicolumn{2}{l}{Xception \cite{rossler2019faceforensics++}} & 86.86$^\ddag$ & 89.30$^\ddag$ &  97.90$^\ddag$ & 99.73$^\ddag$ & \ \ \  \textendash\  & \ \ \ \textendash\  & \ \ \ \textendash\  & \ \ \ \textendash\  \\
    \multicolumn{2}{l}{F$^3$-Net~\cite{qian2020thinking}} & \underline{93.02}$^\dag$ & \underline{95.80}$^\dag$ & 95.97\ \ \   & 98.70\ \ \ &  95.84\ \ \   & 97.51\ \ \   & 76.12\ \ \ & 88.47\ \ \  \\
    \multicolumn{2}{l}{MultiAtt~\cite{zhao2021multi}} & 88.69$^\dag$ & 90.40$^\dag$ & 97.92\ \ \   & \underline{99.94}\ \ \  &  97.26\ \ \ & 99.12\ \ \ & 76.88\ \ \ & 90.14\ \ \ \\
    \multicolumn{2}{l}{PEL~\cite{gu2022exploiting}} & 90.52$^\dag$ & 94.28$^\dag$ & 98.52\ \ \   & 99.63\ \ \  &  97.58\ \ \ & 99.86\ \  & 80.37\ \ \ & 91.06\ \ \ \\
    \multicolumn{2}{l}{RECCE~\cite{cao2022end}} & 91.03$^\dag$ & 95.02$^\dag$ & \underline{98.59}$^\dag$ & \underline{99.94}$^\dag$ &  \underline{97.63}\ \ \ & \underline{99.86}\ \ \ & \underline{81.20}$^\dag$ & \underline{91.33}$^\dag$ \\
    \multicolumn{2}{l}{BENet (ours)} & \textbf{96.83\ \ } & \textbf{98.72\ \ } & \textbf{99.23\ \ } & \textbf{99.98\ \ } &  \textbf{98.96\ \ } & \textbf{99.93\ \ } & \textbf{90.43\ \ } & \textbf{96.38\ \ } \\
    \bottomrule
  \end{tabularx}
  \end{threeparttable}
  \caption{Comparison of BENet intra-dataset performance on Celeb-DF, FF++, DFFD, and DFDC with other state-of-the-arts. $^\dag$ Results taken from their original papers. $^\ddag$ Results taken from RECCE~\cite{cao2022end}.}
  \label{tab:intra}
\end{table*}


To further demonstrate the robustness of BENet more comprehensively, we expand our experimental scope to include the two other versions of FF++, namely FF++ (RAW) and FF++ high-quality (HQ). The latter is also known in the literature as ``C23". The FF++ (HQ) dataset is a well-recognized benchmark, as numerous existing face forgery detection methodologies have been assessed using it. The results of intra-dataset evaluations are shown in \cref{tab:7}. Again, it is clear that BENet not only outperforms other existing approaches but again showcases its consistent capabilities and robustness in the intra-dataset scenario for both uncompressed (RAW) and High-Quality image data.

\begin{table}[htb]
  \centering
  \begin{threeparttable}
  \begin{tabularx}{\columnwidth}{XXXXXX}
    \toprule
   \multicolumn{2}{c}{\multirow{2}{*}{Methods}} & \multicolumn{2}{c}{FF++(RAW)} & \multicolumn{2}{c}{FF++(HQ)}\\ 
   \cline{3-6}
   \multicolumn{2}{c}{\multirow{2}{*}{}} & Acc(\%) & AUC(\%) & Acc(\%) & AUC(\%)  \\
    \midrule
    \multicolumn{2}{l}{Multi-task \cite{nguyen2019multi}} & \ \ \ \textendash & \ \ \ \textendash & 85.65 & 85.43   \\
    \multicolumn{2}{l}{Xception \cite{rossler2019faceforensics++}} & 99.00 & 99.80 & 95.73 & 96.30 \\
    \multicolumn{2}{l}{F$^3$-Net~\cite{qian2020thinking}} & 99.90 & 99.84 & \underline {98.95} & 99.30 \\
    \multicolumn{2}{l}{Face X-ray~\cite{li2020face}} & 99.10 & 99.80 & 97.40 & 97.80 \\
    \multicolumn{2}{l}{MultiAtt~\cite{zhao2021multi}} & \ \ \ \textendash & \ \ \ \textendash & 92.60 & 99.29 \\
    \multicolumn{2}{l}{PEL~\cite{gu2022exploiting}} & \ \ \ \textendash & \ \ \ \textendash & 97.63 & 99.32 \\
    \multicolumn{2}{l}{LipForensics~\cite{haliassos2021lips}} &  98.90 & \underline{99.90} & 98.80 & 99.70 \\
    \multicolumn{2}{l}{RECCE~\cite{cao2022end}} & \ \ \ \textendash & \ \ \ \textendash& 97.06 & 99.32 \\
    \multicolumn{2}{l}{CD-Net~\cite{song2022adaptive}} & \underline{99.91} & \underline
    {99.90} & 98.93 & \underline{99.90} \\
    \multicolumn{2}{l}{BENet (ours)} & \textbf{99.94} & \textbf{99.98} & \textbf{99.29} & \textbf{99.95} \\
    \bottomrule
  \end{tabularx}
  \end{threeparttable}
  \caption{Comparison of BENet intra-dataset performance on RAW and HQ versions of FF++ with other state-of-the-arts. Best results are bolded, runner-up results are underlined.}
  \label{tab:7}
\end{table}



%
\subsubsection{Cross-evaluation} We also conduct a comprehensive cross-dataset evaluation of BENet to demonstrate its superior performance on unseen data when compared to other methods. Here, we use FF++ (LQ) as training data and we test BENet's performance against other models on Celeb-DF, DFFD, and DFDC. 
The results reported in \cref{tab:crossdataLQ} demonstrate the robustness of BENet in cross-dataset evaluation, surpassing its competitors with substantial AUC scores of 77.86\%, 76.59\%, and 78.75\% on Celeb-DF, DFFD, and DFDC, respectively.
More specifically, BENet improves the best AUC in Celeb-DF with 9.15\% achieved by PEL~\cite{gu2022exploiting}. It also improves the best AUC scores achieved by RECCE~\cite{cao2022end} in the DFFD dataset with a large margin of 7.63\%. 
In the challenging DFDC dataset, where existing methods typically show modest performance below the 69.06\% bar, BENet achieves a \textit{state-of-the-art 78.75\% accuracy}, surpassing its closest competitor, RECCE, by 9.69\%. 

\begin{table}
  \centering
  \begin{threeparttable}
  \begin{tabularx}{\columnwidth}{XXXXX}
    \toprule 
    \multicolumn{2}{c}{\multirow{2}{*}{Methods}} & \multicolumn{3}{c}{Test AUC(\%)}\\
    \cline{3-5}
   \multicolumn{2}{c}{\multirow{2}{*}{}} & Celeb-DF & DFFD & DFDC\\ 
    \midrule
    \multicolumn{2}{l}{F$^3$-Net~\cite{qian2020thinking}} & 61.51\ \ & 63.20\ \  & 64.83\ \ \\
    \multicolumn{2}{l}{MultiAtt~\cite{zhao2021multi}} & 67.44$^\dag$ & 67.14\ \ & 68.07\ \ \\
    \multicolumn{2}{l}{PEL~\cite{gu2022exploiting}} & \underline{69.18}$^\dag$ & 66.83\ \  & 63.31$^\dag$\\
    \multicolumn{2}{l}{RECCE~\cite{cao2022end}} & 68.71$^\dag$ & \underline{68.96}\ \  & \underline{69.06}$^\dag$\\
    \multicolumn{2}{l}{BENet (ours)} & \textbf{77.86}\ \  & \textbf{76.59}\ \  & \textbf{78.75}\ \ \\
    
    \bottomrule
  \end{tabularx}
  \end{threeparttable}
  \caption{Comparison of AUC in cross-dataset performance by training on FF++(LQ) with respect to other state-of-the-arts. Best results are bolded, runner-up results are underlined. $^\dag$ Results taken from their original papers.}
  \label{tab:crossdataLQ}
\end{table}

To further validate the robustness of BENet, we train it on FF+ (HQ) and test it on Celeb-DF, DFFD, and DFDC. The results, as shown in \cref{tab:crossdataHQ}, indicate that BENet surpasses competing methods by a considerable margin. For example, in comparison to the next best performer, RECCE, BENet demonstrates a notable improvement with margins of 7.94\% and 6.19\% in the Celeb-DF and DFDC datasets, respectively. These findings again underscore BENet's effectiveness and adaptability in various testing scenarios, reinforcing its potential as a robust solution for face forgery detection.

\begin{table}[htb]
  \centering
  \begin{threeparttable}
  \begin{tabularx}{\columnwidth}{XXXXX}
    \toprule 
    \multicolumn{2}{c}{\multirow{2}{*}{Methods}} & \multicolumn{3}{c}{Test AUC(\%)}\\
    \cline{3-5}
   \multicolumn{2}{c}{\multirow{2}{*}{}} & Celeb-DF & DFFD & DFDC \\ 
    \midrule
    \multicolumn{2}{l}{Xception \cite{rossler2019faceforensics++}}& 65.27 & \ \ \ \textendash & 69.90 \\
    \multicolumn{2}{l}{F$^3$-Net~\cite{qian2020thinking}} & 71.21 & \ \ \  \textendash & 72.88\\
    \multicolumn{2}{l}{Face X-ray~\cite{li2020face}} & 74.20 & \ \ \ \textendash & 70.00 \\
    \multicolumn{2}{l}{RECCE~\cite{cao2022end}} & \underline{77.39} & \ \ \ \textendash &  \underline{76.75}\\
    \multicolumn{2}{l}{BENet (ours)} & \textbf{85.33} & \textbf{81.07} & \textbf{82.94}\\
    \bottomrule
  \end{tabularx}
  \end{threeparttable}
  \caption{Comparison of BENet cross-dataset performance by training on FF++(HQ) with respect to other state-of-the-arts. Best results are bolded, runner-up results are underlined.}
  \label{tab:crossdataHQ}
\end{table}

From the above impressive results on both intra- and cross-evaluations, we believe that this can be attributed to two key factors: 1) BENet's approach of leveraging two modules -Bias expansion and LSA- for expanding ``real-fake" reconstruction as well as learning richer variational relationships of latent-space features across multiple scales between real and fake faces; and more importantly 2) Incorporation of the cross-domain detector during inference to enhance its recognition accuracy.\\

We further offer insights into the effectiveness of face forgery detection methods when initially trained on one manipulation technique and then evaluated on another, following the evaluation protocol used in~\cite{cao2022end}. As shown in \cref{tab:crossmanip}, BENet consistently outperforms other methods across all face manipulation methods, securing the highest AUC scores for each manipulation type. We argue that BENet's success is attributed to its robust feature representation learning strategy, which generalizes well beyond specific training data forgery signatures, enabling effective detection even with new, unseen manipulation methods.

\begin{table}[htb]
  \centering
  \begin{threeparttable}
  \begin{tabularx}{\columnwidth}{XXXXXXX}
    \toprule
    \multirow{2}{*}{Train} & \multicolumn{2}{c}{\multirow{2}{*}{Methods}} & \multicolumn{4}{c}{Test AUC(\%) }\\
    \cline{4-7}
    \multirow{2}{*}{} & \multicolumn{2}{c}{\multirow{2}{*}{}} & DF & FS& F2F& NT \\
    \midrule
    \multirow{4}{*}{DF} & \multicolumn{2}{l}{F$^3$-Net~\cite{qian2020thinking}} & \cellcolor{gray!30} 98.62$^\dag$ & 73.10\ \ & \underline{72.38}\ \  & \underline{70.39}\ \ \\
    \multirow{4}{*}{} & \multicolumn{2}{l}{PEL~\cite{gu2022exploiting}} & \cellcolor{gray!30}99.43\ \  & 70.48\ \  & 68.32\ \  & 67.15\ \ \\
    \multirow{4}{*}{} & \multicolumn{2}{l}{RECCE~\cite{cao2022end}} & \cellcolor{gray!30} \underline{99.65}$^\dag$ & \underline{74.29}$^\dag$ & 70.66$^\dag$ & 67.34$^\dag$\\
    \multirow{4}{*}{} & \multicolumn{2}{l}{BENet (ours)} & \cellcolor{gray!30}\textbf{99.86}\ \  & \textbf{80.75}\ \  & \textbf{78.42}\ \  & \textbf{75.48}\ \ \\
    \midrule
     \multirow{4}{*}{FS} & \multicolumn{2}{l}{F$^3$-Net~\cite{qian2020thinking}} & \underline{83.92}\ \  & \cellcolor{gray!30} 97.23$^\dag$ & 62.89\ \  & 56.28\ \ \\
     \multirow{4}{*}{} & \multicolumn{2}{l}{PEL~\cite{gu2022exploiting}} & 82.01\ \  & \cellcolor{gray!30}97.87\ \  & 62.19\ \  & 50.27\ \ \\
     \multirow{4}{*}{} & \multicolumn{2}{l}{RECCE~\cite{cao2022end}} & 82.39$^\dag$ & \cellcolor{gray!30} \underline{98.82}$^\dag$ & \underline{64.44}$^\dag$ & \underline{56.70}$^\dag$\\
     \multirow{4}{*}{} & \multicolumn{2}{l}{BENet (ours)} & \textbf {86.44}\ \  & \cellcolor{gray!30}\textbf{99.23}\ \  & \textbf{76.28}\ \  & \textbf{75.93}\ \ \\
    \midrule
     \multirow{4}{*}{F2F} & \multicolumn{2}{l}{F$^3$-Net~\cite{qian2020thinking}} & 75.28\ \  & \underline{68.39}\ \  & \cellcolor{gray!30} 95.84$^\dag$ & \underline{72.39}\ \ \\
    \multirow{4}{*}{} &  \multicolumn{2}{l}{PEL~\cite{gu2022exploiting}} & 73.23\ \  & 64.21\ \  & \cellcolor{gray!30}96.38\ \  & 70.96\ \ \\
     \multirow{4}{*}{} & \multicolumn{2}{l}{RECCE~\cite{cao2022end}} & \underline{75.99}$^\dag$ & 64.53$^\dag$ & \cellcolor{gray!30} \underline{98.06}$^\dag$ & 72.32$^\dag$\\
     \multirow{4}{*}{} & \multicolumn{2}{l}{BENet (ours)} & \textbf{82.78}\ \  & \textbf{74.86}\ \  & \cellcolor{gray!30}\textbf{99.08}\ \  & \textbf{76.94}\ \ \\
    \midrule
     \multirow{4}{*}{NT} & \multicolumn{2}{l}{F3-Net~\cite{qian2020thinking}} & \underline{78.83}\ \  & \underline{65.28}\ \  & \underline{83.22}\ \  & \cellcolor{gray!30} 86.01$^\dag$\\
     \multirow{4}{*}{} & \multicolumn{2}{l}{PEL~\cite{gu2022exploiting}} & 72.94\ \  & 60.48\ \  & 72.93\ \  & \cellcolor{gray!30}\underline{94.89}\ \ \\
     \multirow{4}{*}{} & \multicolumn{2}{l}{RECCE~\cite{cao2022end}} & \underline{78.83}$^\dag$ & 63.70$^\dag$ & 80.89$^\dag$ & \cellcolor{gray!30} 94.47$^\dag$\\
     \multirow{4}{*}{} & \multicolumn{2}{l}{BENet (ours)}	 & \textbf{84.63}\ \  & \textbf{77.33}\ \  & \textbf{89.29}\ \  & \cellcolor{gray!30}\textbf{96.84}\ \ \\
    \bottomrule
  \end{tabularx}

  \end{threeparttable}
  \caption{BENet generalization to unseen-manipulations. We report the AUC scores of four manipulation techniques on the FF++ dataset. Close-set evaluations (train and test on the same dataset) are highlighted in gray. Best results are bolded, runner-up results are underlined. $^\dag$ Results taken from their original papers.}
  \label{tab:crossmanip}
\end{table}

\subsubsection{Robustness to Unseen Perturbations} 
\begin{figure*}[htb]
   \centering
   \includegraphics[width=\textwidth]{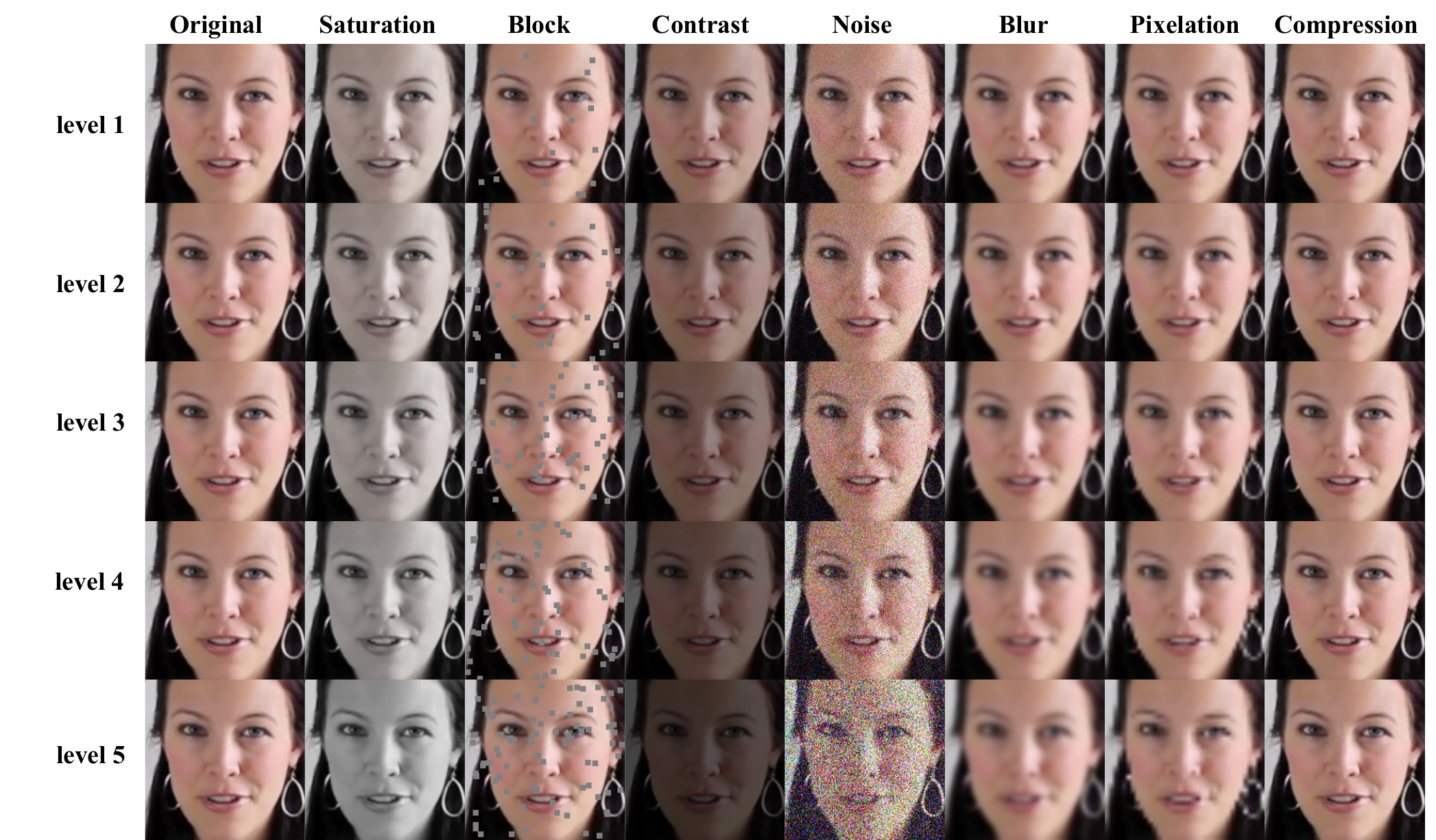}
   \caption{Examples of perturbations at varying levels of severity. These perturbations are introduced in~\cite{jiang2020deeperforensics} and consist of: changes in saturation, adding block-wise distortions, changes in contrast, adding white Gaussian noise, blurring, pixelating, and applying video compression.}
   \label{fig:perturbations}
\end{figure*}

In the current landscape of social media, where image processing is widespread, it is essential that forgery detection systems remain effective against common image perturbations~\cite{haliassos2021lips}. In this section, we conduct novel experiments to demonstrate the robustness of BENet to unseen perturbations. We train our model on FF++ (RAW) and test it on samples from the same dataset that had been subjected to various unseen distortions. Following the methodologies used in previous studies~\cite{jiang2020deeperforensics,haliassos2021lips}, we create seven different types of perturbations, each at five levels of severity, as shown in \cref{fig:perturbations}.  To achieve this, we use the DeeperForensics code\footnote{https://github.com/EndlessSora/DeeperForensics-1.0/tree/master/perturbation}.

\begin{figure*}[htb]
   \centering
   \includegraphics[width=\textwidth]{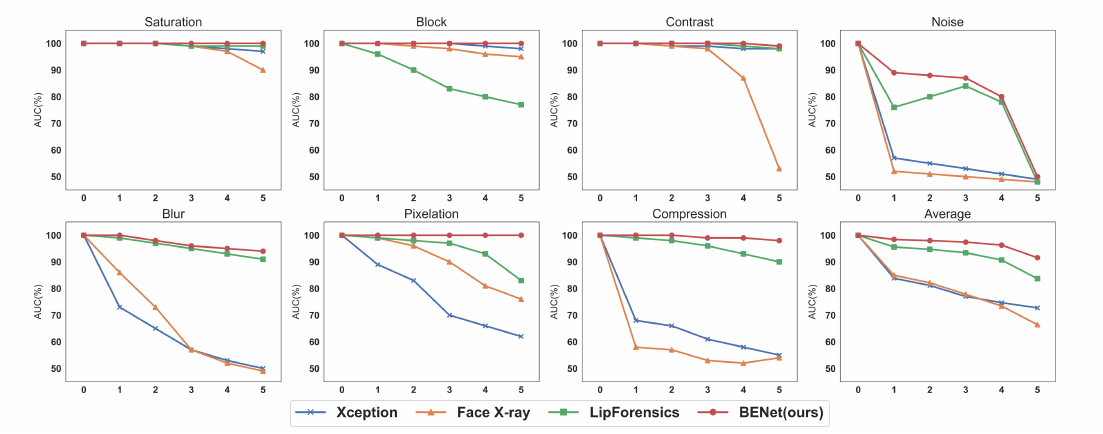}
   \caption{Robustness to various perturbation examples of different severity levels."Average" denotes the mean across all perturbations at each severity level. BENet is more robust than state-of-the-art approaches to all types of perturbations.}
   \label{fig:RobPerturb}
\end{figure*}

\cref{fig:RobPerturb} illustrates the robustness of models to unseen perturbations with increasing severity levels. We compare BENet with three contemporary approaches, namely Face X-ray~\cite{li2020face}, Xception \cite{rossler2019faceforensics++}, and LipForensics~\cite{haliassos2021lips}.   We observe a steady robustness in our approach against all various types of perturbations. On average, BENet achieves far better performance compared to other methods. Specifically, when faced with perturbations that impact the high-frequency clues of video frames, like blurring, pixelation, or compression, BENet retains robust performance across all but the most severe levels. In contrast, other methods show a marked drop in their effectiveness under these conditions. Although Face X-ray exhibits strong performance across various cases, it is notably vulnerable to most perturbations, particularly compression, indicating a high susceptibility of the blending boundary to corruption ~\cite{haliassos2021lips}. Hence, our approach excels in robustness thanks to three key innovations: the Bias Expansion module preserves authentic features while emphasizing fake discrepancies, our Latent-space Attention (LSA) module captures crucial forgery-related inconsistencies, and the Cross-Domain Detector module enhances recognition accuracy, particularly in facial domain verification. These innovations collectively make our approach a formidable solution for the task.

In \cref{tab:AUCDecay}, we present the \textbf{decay of AUC scores} for each perturbation across all severity levels. The results indicate that BENet exhibits the least degradation in all cases. This robustness is primarily attributed to the powerful modules we integrate within BENet, namely Bias expansion,Latent-space attention, and Cross-domain detector.

\begin{table*}
  \centering
  \begin{tabularx}{\textwidth}{XXXXXXXXXXX}
    \toprule
    \multicolumn{2}{l}{Methods} & Saturation & Block & Contrast & Noise & Blur & Pixelation & \multicolumn{2}{c}{Compression} & Average\\
    \midrule
    \multicolumn{2}{l}{Xception \cite{rossler2019faceforensics++}} & -0.5 & -0.1 & -1.2 & -46.0 & -39.6 & -25.6 & \multicolumn{2}{l}{-37.7} & -21.5\\
    \multicolumn{2}{l}{Face X-ray~\cite{li2020face}} & -2.2 & -0.6 & -11.3 & -50.0 & -36.0 & -11.2 & \multicolumn{2}{l}{-44.6} & -22.3 \\
    \multicolumn{2}{l}{LipForensics~\cite{haliassos2021lips}} &-0.0 & -12.5 & -0.3 & -26.1 & -3.8 & -4.3 & \multicolumn{2}{l}{-4.3} & -7.4\\
    \multicolumn{2}{l}{BENet (ours)} & \textbf{-0.0} & \textbf{-0.0} & \textbf{-0.2} & \textbf{-17.7} & \textbf{-2.81} & \textbf{-0.1} & \multicolumn{2}{l}{\textbf{-0.7}} &  \textbf{-3.1}\\
    \bottomrule
  \end{tabularx}
  \caption{Evaluating BENet's Robustness on FF++ (RAW)~\cite{rossler2019faceforensics++}  through AUC (\%) decay under perturbations. We report the change in the video constant rate factor. BENet outperforms other state-of-the-arts and shows superior robustness to unseen perturbations.}
  \label{tab:AUCDecay}
\end{table*}

\subsection{Ablation Study}
\label{sec:5.3}
In this section, we conduct exhaustive ablations to evaluate the effectiveness of various components and training strategies within BENet. We systematically modify the network by adding or removing specific elements, enabling a detailed analysis of each component's contribution, as depicted in \cref{tab:4}. Our baseline is the Xception model, a standard classification approach \cite{rossler2019faceforensics++}. The training strategies  include different loss functions: $L_{c}$, $L_{1}$, $L_{be}$, and $L_{rec}$, with the latter being the vanilla \textit{mean squared error (MSE)} used for reconstruction purposes. Notably, in variation (\textcircled{\scriptsize 5}), the reconstruction learning is applied exclusively to real faces, similar to the approach used in RECCE \cite{cao2022end}.

\begin{table}
  \centering
  \begin{tabularx}{\columnwidth}{XXXXX}
    \toprule
    \multicolumn{3}{l}{Methods} & Acc(\%) & AUC(\%) \\
    \midrule
    \multicolumn{3}{l}{\textcircled{\scriptsize 1} Baseline($L_{c}$)} & 82.43 & 86.67 \\
    \multicolumn{3}{l}{\textcircled{\scriptsize 2} AE($L_{rec}$ + $L_{c}$)} & 84.38 & 88.43 \\
     \multicolumn{3}{l}{\textcircled{\scriptsize 3} AE+Bias($L_{rec}$ + $L_{c}$)} & 85.46 & 90.30\\
    \multicolumn{3}{l}{\textcircled{\scriptsize 4} AE+Bias+LSA($L_{rec}$ + $L_{c}$)} & 87.34 & 92.07 \\
    \multicolumn{3}{l}{\textcircled{\scriptsize 5} AE+Bias+LSA($L_{1}$ + $L_{c}$)} & 89.67 & 94.79 \\
    \multicolumn{3}{l}{\textcircled{\scriptsize 6} AE+Bias+LSA+CD($L_{1}$ + $L_{c}$)} & 92.25 & 96.71 \\
    \multicolumn{3}{l}{\textcircled{\scriptsize 7} BENet w/o CD($L_{be}$ + $L_{c}$)} & 93.41 & 96.33 \\
    
   \multicolumn{3}{l}{Full BENet($L_{be}$ + $L_{c}$)}  & \textbf{96.83} & \textbf{98.72} \\
    \bottomrule
  \end{tabularx}
  \caption{Ablation study on FF++(LQ) with different components and training strategies. AE: Autoencoder; CD: Cross-domain detector.}
  \label{tab:4}
\end{table}


\subsubsection{Effectiveness of bias calculation} The inclusion of an auto-encoder (\textcircled{\scriptsize 2}) for reconstructing input face images significantly enhances performance, with a 1.95\% increase in Acc and 1.76\% in AUC compared to the baseline variation (\textcircled{\scriptsize 1}). Further improvements are observed when bias images are calculated (\textcircled{\scriptsize 3}), leading to an additional 1.08\% rise in Acc and 1.87\% in AUC. These results suggest that using autoencoder reconstructions in BENet to amplify deepfake clues is effective, and the computation of bias images further aids in making this information more straightforward for network optimization.

\subsubsection{Effectiveness of bias expansion loss} As previously mentioned, the bias expansion loss $L_{be}$ plays a pivotal role in guiding BENet's learning process to discern the bias within the reconstruction of real and fake faces. Thus, its formulation encompasses two elements: An \textit{invariant} reconstruction component for real faces and a \textit{bias} component for fake faces. Compared to variations without $L_{be}$ (\textcircled{\scriptsize 5}, and \textcircled{\scriptsize 6}), i.e., those solely utilizing reconstruction loss for real faces ($L_{1}$), the integration of bias expansion loss (\textcircled{\scriptsize 7}) results in a significant improvement in both Acc and AUC.

\subsubsection{Effectiveness of latent-space attention} Excluding the LSA  from BENet leads to a reduced ability to detect inconsistencies in subtle forgeries within the latent space. Compared to (\textcircled{\scriptsize 3}), adding the LSA module (\textcircled{\scriptsize 4}) brings an improvement of 1.88\% and 1.77\% in Acc and AUC, respectively. LSA lets BENet emphasize the forgery-related inconsistencies by capturing the latent feature variances across different scales for improved fine-grained manipulation detections. 

\subsubsection{Effectiveness of cross-domain detector} We also examine the role of the cross-domain detector in BENet. When it is omitted (\textcircled{\scriptsize 7}), there is a significant drop in the model's ability to handle cross-domain forgeries. As such, We gain a fair improvement of 3.42\% and 2.39\% in Acc and AUC, respectively, with the full BENet model. This proves that cross-domain detector is instrumental for improved generalization of BENet on unseen cross-domain fake faces.  


\subsubsection{Impact of hyper-parameter \textbf{$\lambda$}} To balance the contribution of $L_c$ and $L_{be}$ in the total loss function, we experiment with various values of $\lambda$ ranging from 0.1 to 1.0 in increments of 0.1. As shown in \cref{tab:5}, BENet maintains the highest performance with $\lambda$ set at 0.5, effectively balancing cross-entropy loss and bias expansion loss. Deviating from this optimal value leads to some compromise in the performance.  Values below 0.5 caused the model to overly prioritise bias expansion, increasing forgery detection while raising considerable false positives. In contrast, values above 0.5 make BENet lean more towards cross-entropy loss, resulting in a more cautious approach to forgery detection. Thus, we stick to  $\lambda=0.5$ as the optimal value.

\begin{table}
  \centering
 \begin{tabularx}{\columnwidth}{XXX}
    \toprule
    $\lambda$ & Acc(\%)  & AUC(\%)  \\
    \midrule
    0.1 & 92.05 & 95.64 \\
  0.2 & 93.64 & 96.33 \\
  0.3 & 95.21 & 96.70\\
   0.4 & 96.34 & 97.46 \\
   \textbf {0.5} & \textbf{96.83} & \textbf{98.72}\\
  0.6 & 96.27 & 98.21 \\
   0.7 & 95.85 & 98.01 \\
   0.8 & 95.13 & 97.52 \\
   0.9 & 94.55 & 97.54\\
  1.0 & 93.25 & 96.71 \\
   \bottomrule
  \end{tabularx}
 \caption{Ablation study on hyper-parameter $\lambda$ of the total loss.}
  \label{tab:5}
\end{table}


\subsubsection{BENet generalization to unseen manipulations within FF++(LQ)} To further consolidate our findings and enrich our ablations, we extend our experimentation by examining how well each component of BENet generalizes to unseen manipulations within FF++(LQ). This is particularly crucial in the context of Deepfake detection, where the ability to adapt to novel and unfamiliar manipulations is a key measure of a model's robustness. In our extended study, we leverage the four distinct manipulation techniques presented in the FF++(LQ), namely DF, FS, F2F, and NT. Our methodology involves a cross-manipulation training and testing approach where we train each component on images manipulated by one specific technique and then test it on images altered using a different manipulation technique. This cross-evaluation strategy is instrumental in determining the adaptability and resilience of each component of BENet when faced with various Deepfake generation methods. The results are shown in Table \cref{tab:6}.

\begin{table}[htb]
  \centering
  \begin{threeparttable}
  \begin{tabular}{@{}lccccccc@{}}
    \toprule
    \multirow{2}{*}{Train} & \multirow{2}{*}{Methods} & \multicolumn{4}{c}{Test AUC(\%)}\\
    \cline{3-6}
    \multirow{2}{*}{} & \multirow{2}{*}{} & DF & FS& F2F& NT \\
    \midrule
    \multirow{8}{*}{DF} & \textcircled{\scriptsize 1} & \cellcolor{gray!30} 86.48 & 56.82 & 54.74 & 50.22\\ 
    \multirow{8}{*}{} & \textcircled{\scriptsize 2} & \cellcolor{gray!30} 87.70 &  57.38 &  55.33 &  51.28\\
    \multirow{8}{*}{} & \textcircled{\scriptsize 3} & \cellcolor{gray!30} 88.54 & 58.32 & 56.46 & 52.49\\
    \multirow{8}{*}{} & \textcircled{\scriptsize 4} & \cellcolor{gray!30} 90.62 & 59.44 & 58.35 & 55.24\\
    \multirow{8}{*}{} & \textcircled{\scriptsize 5} & \cellcolor{gray!30} 92.25 & 66.92 & 65.28 & 62.93\\
    \multirow{8}{*}{} & \textcircled{\scriptsize 6} & \cellcolor{gray!30} 95.74 & 75.28 & 73.48 & 69.45\\
    \multirow{8}{*}{} & \textcircled{\scriptsize 7} & \cellcolor{gray!30} 96.43 & 73.24 & 72.91 & 68.36\\
   \multirow{8}{*}{} & Full BENet & \cellcolor{gray!30} \textbf{99.86} & \textbf{80.75} & \textbf{78.42} & \textbf{75.48}\\
   \midrule
   \multirow{8}{*}{FS} & \textcircled{\scriptsize 1} & 63.28 & \cellcolor{gray!30} 85.79 & 57.83 & 56.33\\
   \multirow{8}{*}{} & \textcircled{\scriptsize 2} & 64.50 & \cellcolor{gray!30} 86.83 & 59.22 & 58.49\\
   \multirow{8}{*}{} & \textcircled{\scriptsize 3} & 65.93 & \cellcolor{gray!30} 87.58 & 61.57 & 60.45\\
   \multirow{8}{*}{} & \textcircled{\scriptsize 4} & 68.39 & \cellcolor{gray!30} 89.92 & 63.92 & 62.86\\
   \multirow{8}{*}{} & \textcircled{\scriptsize 5} & 73.58 & \cellcolor{gray!30} 92.23 & 68.34 & 66.20\\
   \multirow{8}{*}{} & \textcircled{\scriptsize 6}  & 80.20 & \cellcolor{gray!30} 95.62 & 73.02 & 71.29\\
   \multirow{8}{*}{} & \textcircled{\scriptsize 7} & 79.24 & \cellcolor{gray!30} 96.08 & 72.55 & 70.32\\
   \multirow{8}{*}{} & Full BENet & \textbf{86.44} & \cellcolor{gray!30} \textbf{99.23} & \textbf{76.28} & \textbf{75.93}\\
   \midrule
   \multirow{8}{*}{F2F} & \textcircled{\scriptsize 1} & 58.13 & 54.25 & \cellcolor{gray!30} 84.98 & 56.28\\
   \multirow{8}{*}{} & \textcircled{\scriptsize 2} & 60.48 & 56.34 & \cellcolor{gray!30} 86.32 & 58.49\\
   \multirow{8}{*}{} & \textcircled{\scriptsize 3} & 62.74 & 58.82 & \cellcolor{gray!30} 87.64 & 59.97\\
   \multirow{8}{*}{} & \textcircled{\scriptsize 4} & 64.92 & 60.38 & \cellcolor{gray!30} 88.92 & 61.45\\
   \multirow{8}{*}{} & \textcircled{\scriptsize 5} & 72.89 & 66.82 & \cellcolor{gray!30} 91.93 & 68.37\\
   \multirow{8}{*}{} & \textcircled{\scriptsize 6} & 77.43 & 70.32	 & \cellcolor{gray!30} 94.74 & 73.81\\
   \multirow{8}{*}{} & \textcircled{\scriptsize 7} & 78.26 & 70.94 & \cellcolor{gray!30} 95.89 & 72.93\\
   \multirow{8}{*}{} & Full BENet & \textbf{82.78} & \textbf{74.86} & \cellcolor{gray!30} \textbf{99.08} & \textbf{76.94}\\
   \midrule
   \multirow{8}{*}{NT} & \textcircled{\scriptsize 1} & 60.27 & 54.89 & 63.12 & \cellcolor{gray!30} 82.39\\
   \multirow{8}{*}{} & \textcircled{\scriptsize 2} & 61.28 & 56.32 & 65.43 & \cellcolor{gray!30} 83.62\\
   \multirow{8}{*}{} & \textcircled{\scriptsize 3} & 63.84 & 58.56 & 67.16 & \cellcolor{gray!30} 85.21\\
   \multirow{8}{*}{} & \textcircled{\scriptsize 4} & 65.21 & 60.39 & 69.42 & \cellcolor{gray!30} 88.48\\
   \multirow{8}{*}{} & \textcircled{\scriptsize 5} & 71.55 & 64.31 & 75.93 & \cellcolor{gray!30} 90.23\\
   \multirow{8}{*}{} & \textcircled{\scriptsize 6} & 78.27 & 70.28 & 82.94 & \cellcolor{gray!30} 92.03\\
   \multirow{8}{*}{} & \textcircled{\scriptsize 7} & 77.48 & 69.32 & 83.01 & \cellcolor{gray!30} 93.41\\
   \multirow{8}{*}{} & Full BENet & \textbf{84.63} & \textbf{77.33} & \textbf{89.29} & \cellcolor{gray!30} \textbf{96.84}\\
    \bottomrule
  \end{tabular}
  \end{threeparttable}
  \caption{Additional ablation study evaluating BENet's generalization to unseen manipulations in the FF++(LQ) dataset, exploring various components and training methods. Evaluations conducted on a closed set (training and testing on the same dataset) are highlighted in gray.}

  \label{tab:6}
\end{table}

Comparing with the ``Full BENet", the following observations can be made: 1) The improvement trend in AUC persists consistently across all cross-manipulations from variant 1 to 7;  2) The Full BENet consistently exhibits significantly better performance compared to other variants.\\
These outcomes provide additional validation for BENet's generalization capacities, originating from its proposed methodological construction. This level of detailed analysis we conducted is crucial in demonstrating the robustness and versatility of BENet in the rapidly evolving domain of Deepfake detection.

\subsection{Experimental and Qualitative Analysis}


\subsubsection{Feature Distribution Analysis} 
\begin{figure}[htb]
   \centering
   \includegraphics[width=\linewidth]{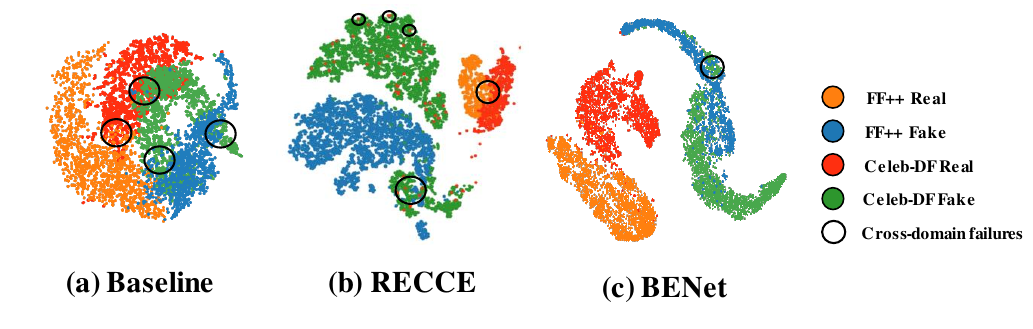}
   \caption{Comparative t-SNE~\cite{van2008visualizing} visualization of feature space separation for real and fake data points across different detection methods.}
   \label{fig:1}
\end{figure}

The learned feature visualizations for the: \textbf{(a)} baseline, \textbf{(b)} RECCE, and \textbf{(c)} BENet, all trained on FF+ (LQ), are presented in \cref{fig:1}. For instance, we extract the features from the penultimate fully-connected layer, selecting 2000 samples each from FF++ for in-domain evaluation and Celeb-DF for cross-domain evaluation. 
The t-distributed stochastic neighbor embedding (t-SNE)~\cite{van2008visualizing} visualizations reveal that BENet learns highly discriminative features. In the t-SNE space, real and fake face embeddings are distinctly separated, more than in the baseline and RECCE. 
This notable separation observed in BENet feature clusters affirms its impressive ability to discern between authentic and counterfeit faces.


\subsubsection{Analysis of the Learned Feature Maps} 
\begin{figure}[htb]
  \centering
   \includegraphics[width=\linewidth]{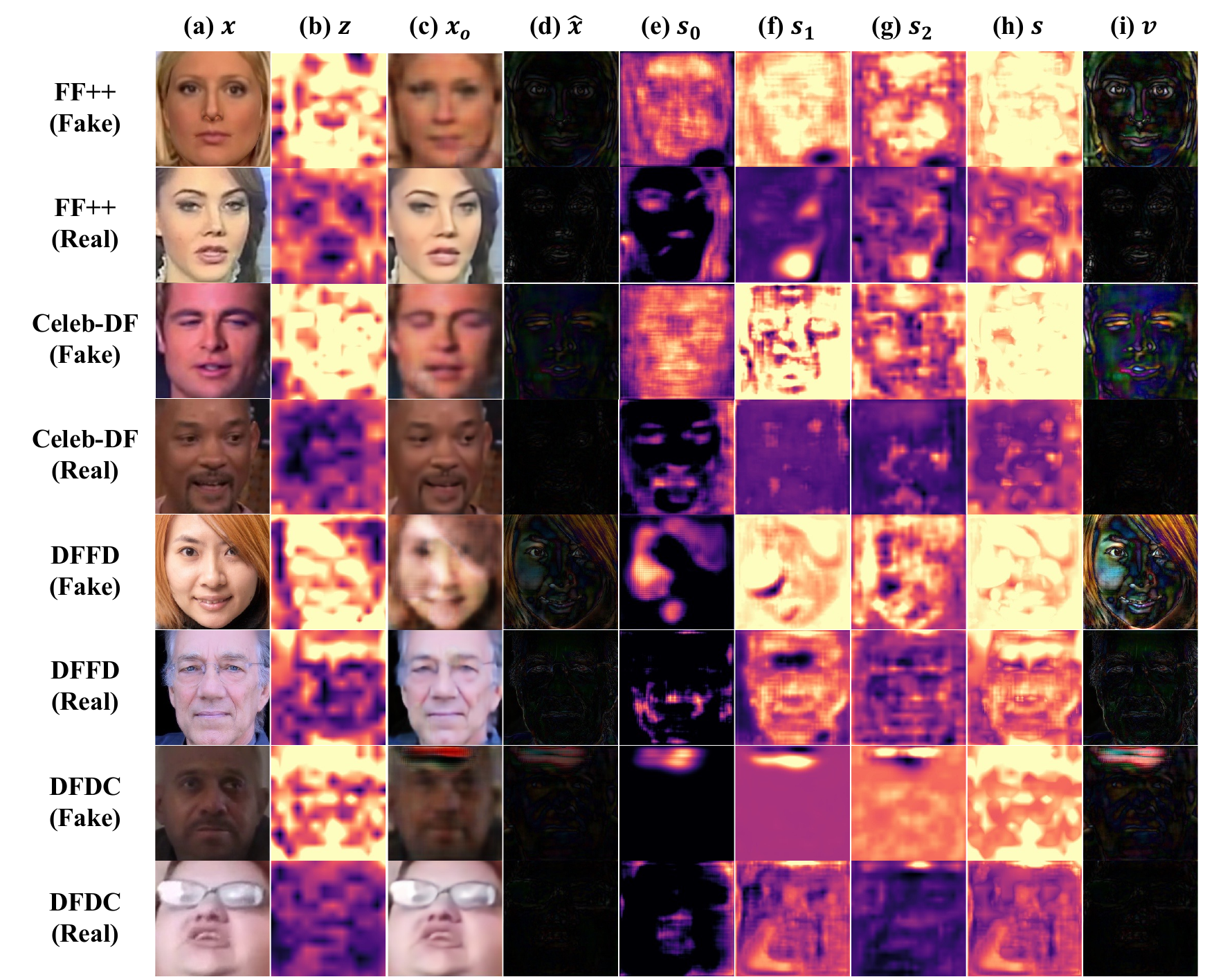}
   \caption{Examplar of real/fake faces, their reconstructions,  and their related feature maps at different levels in BENet,  and across various datasets.}
   \label{fig:features}
\end{figure}

We visualize the real/fake faces reconstructions and their related learned feature maps at different levels in BENet across various datasets. The results are shown in \cref{fig:features}. We  observe the progressive refinement and adeptness of BENet in isolating and accentuating forgery clues across fake faces. First, we can notice the ability of BENet to reconstruct real faces consistently, with a little blur, while adaptively amplifying the deepfake clues present in fake faces (c). More specifically, The LSA module finely tunes its focus through the attentive examination of multi-scale representations derived from the bias expansion, subsequently intensifying the anomalies within the Bias image. Large-scale feature maps offer extensive but noisy clues, while small-scale ones provide detailed yet partial clues. Since the autoencoder differentially reconstructs to discover forgeries, the LSA module strives to further enhance them by capturing the differences in multi-scale features. Thus, for real faces characterized by consistent reconstruction, the attention map $s$ holds a minimal weight, resulting in a negligible difference in the composite image $v$, \textit{i.e.,} often manifesting as a near-black image. Conversely, with fake faces, the reconstruction discrepancies elevate the $s$ map's weight, and $v$ accentuates the divergence rooted in $\hat{x}$. LSA further sharpens the classifier's ability to differentiate and flag deepfakes with greater capabilities by expanding the gap between real and fake images.

\subsubsection{Decision-making Analysis} 
\begin{figure}[htb]
  \centering
   \includegraphics[width=\linewidth]{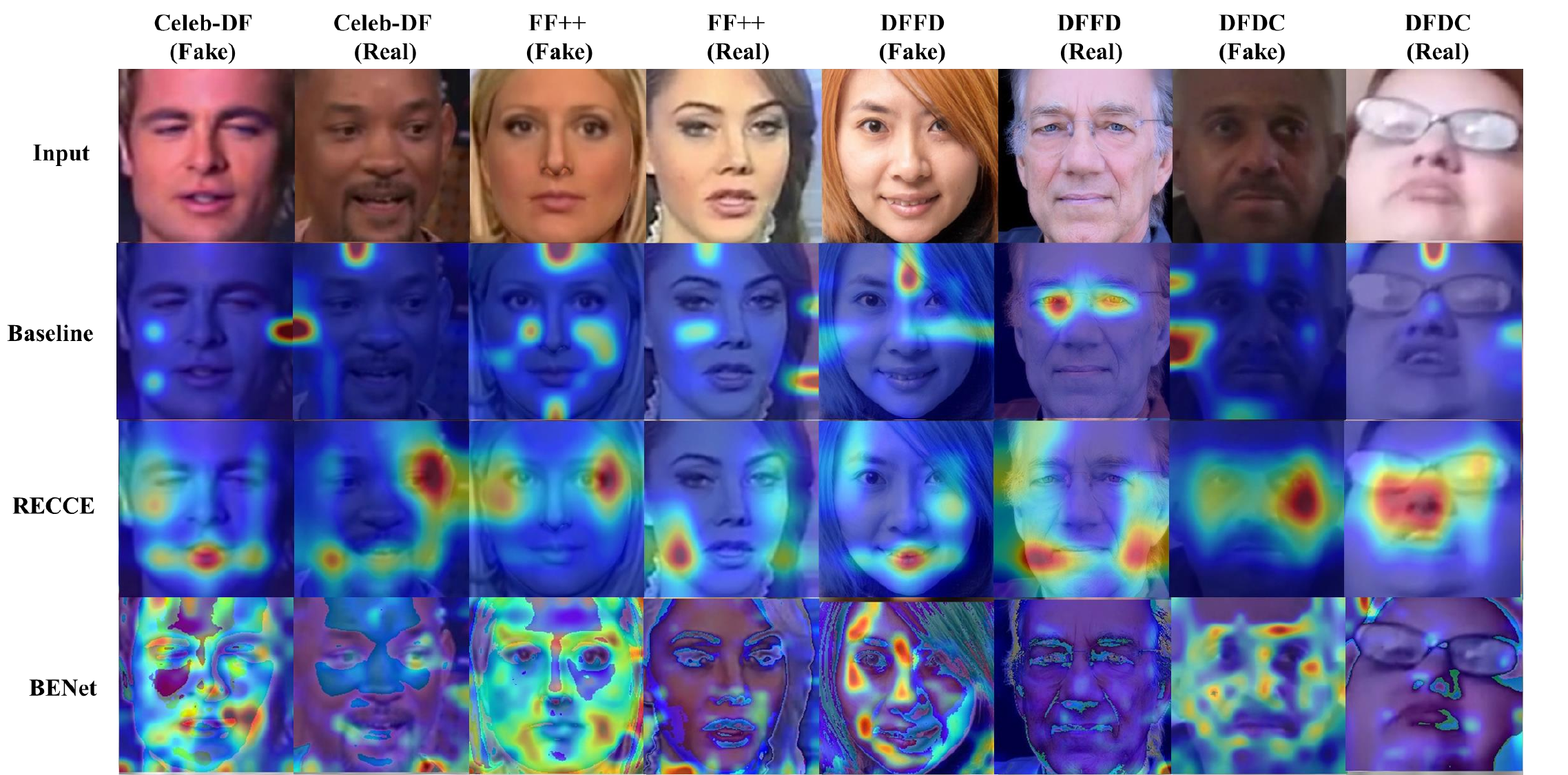}
   \caption{The Grad-CAM~\cite{zhou2016learning} visualizations with different models across various datasets.}
   \label{fig:5}
\end{figure}

\cref{fig:5} visualizes the gradient-weighted  activation mapping (Grad-CAM)~\cite{zhou2016learning} of our method in different evaluation settings. The comparative analysis of Grad-CAM outputs across all datasets shows that BENet has a clear advantage when detecting deepfake manipulations. In contrast to other models, BENet reliably produces heatmaps with distinct, well-defined areas of emphasis, especially over important face characteristics where forgeries are most detectable (e.g., eye, nose, and mouth). On the flip side, it generates fewer heatmaps on the real faces, as opposed to other methods. We argue that the multi-scale LSA module within BENet is crucial not only for highlighting subtle inconsistencies that other models overlook but also adapts to the holistic context of the face, ensuring a robust defense against even aginst the most sophisticated forgeries. 

\section{Limitations of the Proposed Method}
\label{sec:limitation}

\begin{figure}[htb]
   \centering
   \includegraphics[width=\columnwidth]{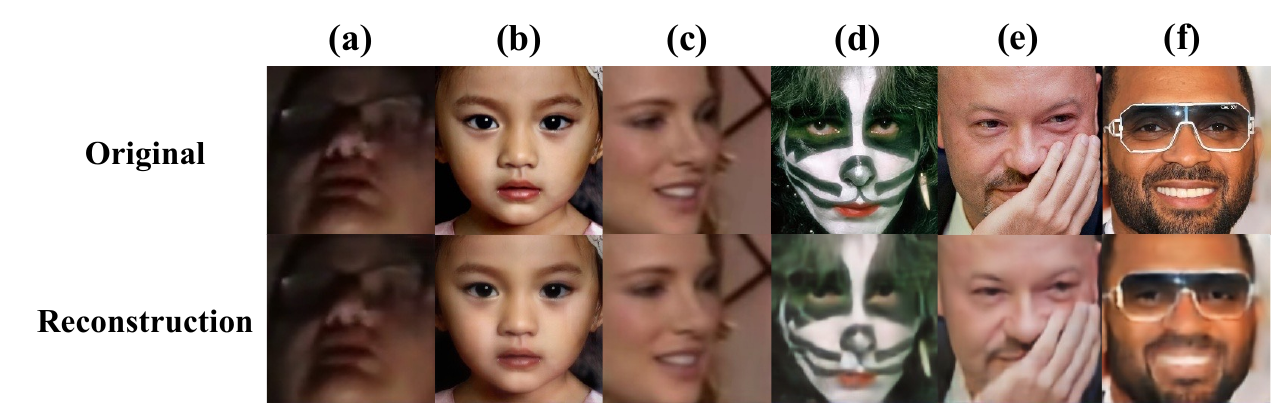}
   \caption{Failure cases visualization. (a) \& (b) \& (c) are the false negative cases (the label is fake, while the prediction is real). (d) \& (e) \& (f) are the false positive cases (the label is real, while the prediction is fake). }
   \label{fig:limitations}
\end{figure}

While BENet demonstrates notable robustness against cross-domain attacks, it may face difficulties when confronted with samples that closely resemble real faces. While it prioritizes the detection of cross-domain deepfakes to streamline incremental learning costs, addressing the nuances of deepfake variations closely mimicking real faces is paramount for enhancing the security of face recognition systems. In \cref{fig:limitations}, we present examples of failure cases encountered by BENet. 

\subsection{False Negative (Predict Fake as Real, Examples a-c)} This category typically includes challenging samples that the model may mistakenly identify as real. These difficulties arise from various factors, such as (1) low-quality images characterized by extreme dimness or poor resolution, (2) subtle or less apparent fake clues, and (3) limited visibility of facial features due to significant posing or angles, which can also be deemed as low-quality traits. These complex samples hinder the model's ability to perform nuanced reconstruction, leading to potential confusion with real images. These issues warrant further investigation in future studies.

\subsection{False Positive (Predict Real as Fake, Examples d-f)} In this category, samples are often erroneously classified as fake faces by the model due to elements like obstructions or alterations. Instances include (d) faces with graffiti, (e) faces partially covered by hands, and (f) faces obscured by sunglasses. These kinds of occlusions can mislead the model into detecting them as artificial clues, thereby resulting in misidentification. This phenomenon highlights the need for models to better discern between genuine occlusions and deliberate falsifications.

Finally, we emphasize the importance of the cross-domain detector erring on the side of false positives rather than false negatives when detecting face forgeries in the provided posterior probability distribution (PPD). False positives, while inconvenient, allow for a secondary verification process. In contrast, false negatives pose a significant threat to security systems, as they represent undetected forgeries. Therefore, while both errors are undesirable, false positives are preferable due to the additional safety net they provide through subsequent verification.

\section{Conclusion}
We proposed BENet, a Cross-Domain Robust Bias Expansion Network for efficient face forgery detection. BENet leverages a bias expansion-based autoencoder to expand bias between real and fake face reconstructions to unveil hidden forged clues. To further tackle subtle forgeries, we incorporated an innovative LSA module designed to capture variations in latent-space between the encoder and decoder. Furthermore, to refine detection outcomes for unfamiliar cross-domain deepfakes, BENet integrates a cross-domain detector activated during inference, significantly boosting recognition accuracy. Rigorous evaluations, both quantitative and qualitative, conducted on SOTA benchmarks, evidenced the robustness of BENet across intra and cross-domain scenarios.

 \section{References Section}
\bibliographystyle{IEEEtran}
\bibliography{main}
\vfill

\end{document}